\RequirePackage{fix-cm}
\documentclass[smallextended]{svjour3_MODIFIED}       
\smartqed  
\usepackage{graphicx}
\journalname{Research in the Mathematical Sciences}

\usepackage{pgfplots}
\usepackage{tikz}
\usepackage{multirow}
\usepackage{amsmath, amssymb}
\usepackage{bm}
\usepackage{makecell}
\usepackage{pgfplotstable}
\usepackage{pdflscape}
\usepackage{float}
\usepackage{afterpage}
\usepackage{bigints}

\usetikzlibrary{matrix,calc,shapes,positioning,pgfplots.fillbetween,backgrounds}

\newcommand\TstrutS{\rule{0pt}{4ex}}        
\newcommand\BstrutS{\rule[-2.4ex]{0pt}{0pt}}

\newcommand\TstrutL{\rule{0pt}{6ex}}        
\newcommand\BstrutL{\rule[-4.4ex]{0pt}{0pt}}

\newcommand{\grad}{\bm \nabla}
\newcommand{\gradT}{\bm \nabla^\top \!}

\tikzset
{
  block/.style = {shape=rectangle, rounded corners,
                  draw, anchor=center, minimum height = 0em,
                  align=center, minimum width = 2em,
                  inner sep=1ex, fill=white},
  operator/.style = {block, circle, inner sep = .3ex, minimum width = 0em, 
                     minimum height = 0em},
  summary/.style = {block, fill=black!10, rounded corners=0cm, minimum height = 
  0em,}
}

\newcommand{\edge}[2]{\draw (#1) edge node (TMP) {} (#2);}

\newcommand{\coupleedgein}[2]{
  \node (nl)  [below left=-2.1mm and 2.5mm of #1.south] {};
  \node (nr)  [below right=-2.1mm and 2.5mm of #1.south] {};
  \node (tl)  [left =0mm of #2.center] {};
  \node (tr)  [right =0mm of #2.center] {};
  \begin{pgfonlayer}{bg1}
  \draw[-] (nl.center) -- (tl.center);
  \draw[-] (nr.center) -- (tr.center);
  \end{pgfonlayer}
  \begin{pgfonlayer}{bg0}
  \fill[black!20] (tr.center) -- (tl.center) -- (nl.center) -- (nr.center) --  
  cycle;
  \end{pgfonlayer}
}

\newcommand{\coupleedgeout}[2]{
  \node (nl)  [above left=-2.1mm and 2.5mm of #2.north] {};
  \node (nr)  [above right=-2.1mm and 2.5mm of #2.north] {};
  \node (tl)  [left =0mm of #1.center] {};
  \node (tr)  [right =0mm of #1.center] {};
  \begin{pgfonlayer}{bg1}
  \draw[-] (tl.center) -- (nl.center);
  \draw[-] (tr.center) -- (nr.center);
  \end{pgfonlayer}
  \begin{pgfonlayer}{bg0}
  \fill[black!20] (nl.center) -- (nr.center) -- (tr.center) -- (tl.center) --  
  cycle;
  \end{pgfonlayer}
}

\newcommand{\layername}[4]{\path (#1) -- node (TMP) {} (#2);
                           \draw node [left of=TMP, xshift=#4] {#3};}

\newcommand{\layernamesplit}[4]{\path (#1) -- +(0,-0.6) -| 
                                 node [near end](TMP) {} (#2);
                               \draw node [left of=TMP, xshift=#4] {#3};}
                             
\newcommand{\edgelabel}[1]{\draw node [right of=TMP, anchor=west, 
                           xshift=-2.2em] {#1};}

\newcommand{\skipedge}[3] {\draw (#1) -- +(#3,0) |-
                           node (TMP) {}
                           node [near start,left] {Id} (#2);}

\newcommand{\splitedge}[2]{\draw (#1) -- +(0,-0.6) -| node [near end] (TMP) {} 
                          (#2);}
                         
\newcommand{\joinedge}[2]{\draw [latex-] (#2) -- +(0,0.6) -| node [near end] 
                          (TMP) {} (#1);}

\spdefaulttheorem{designprinciple}{Design Principle}{\bfseries}{\itshape}

\begin{document}
\title{Designing Rotationally Invariant Neural Networks from PDEs and 
Variational Methods
\thanks{This work has received funding from the European Research Council (ERC) 
under the European Union's Horizon 2020 research and innovation programme 
(grant agreement no. 741215, ERC Advanced Grant INCOVID).}
}
\subtitle{}

\titlerunning{Designing Rotationally Invariant Networks from PDEs and 
Variational Methods}        

\author{Tobias Alt \and Karl Schrader \and \\Joachim Weickert \and Pascal Peter 
\and Matthias Augustin}


\institute{All authors are with the
Mathematical Image Analysis Group,
Faculty of Mathematics and Computer Science,
Campus E1.7, Saarland University,
66041 Saarbr\"ucken, Germany.\\
\email{alt@mia.uni-saarland.de}
}

\date{Received: date / Accepted: date}

\maketitle

\begin{abstract}
Partial differential equation (PDE) models and their associated variational 
energy formulations are often rotationally invariant by design. This ensures 
that a rotation of the input results in a corresponding rotation of the output, 
which is desirable in applications such as image analysis.
Convolutional neural networks (CNNs) do not share this property, and existing 
remedies are often complex. The goal of our paper is to investigate how 
diffusion and variational models achieve rotation invariance and transfer 
these ideas to neural networks. As a core novelty we propose activation 
functions which couple network channels by combining information from several 
oriented filters. This guarantees rotation invariance within the basic 
building blocks of the networks while still allowing for directional filtering. 
The resulting neural architectures are inherently rotationally invariant. With 
only a few small filters, they can achieve the same invariance as existing 
techniques which require a fine-grained sampling of orientations. Our findings 
help to translate diffusion and variational models into mathematically 
well-founded network architectures, and provide novel concepts for model-based 
CNN design.

\end{abstract}

\section{Introduction} 
Partial differential equations (PDEs) and variational methods are core 
parts of various successful model-based image processing approaches; see 
e.g.~\cite{AK06,CS05a,We97} and the references therein. Such models often 
achieve invariance under transformations such as translations and 
rotations by design. These invariances reflect the physical motivation of the 
models: Transforming the input should lead to an equally transformed output. 

Convolutional neural networks (CNNs) and deep learning 
\cite{GBC16,LBH15,LBBH98,Sch15a} have revolutionised the field of image 
processing in recent years. The flexibility of CNN models allows to apply them 
to various tasks in a plug-and-play fashion with remarkable performance. Due to 
their convolution structure, CNNs are shift invariant by design. However, they  
lack inherent rotation invariance. Proposed adaptations often inflate the 
network structure and rely on complex filter design with large stencils; see 
e.g.~\cite{WHS18}.

In the present paper, we tackle these problems by translating 
rotationally invariant PDEs and their corresponding variational formulations 
into neural networks. This alternative view on rotation invariance within 
neural architectures yields novel design concepts which have not yet been 
explored in CNNs. 

Since in the literature, multiple notions of rotation invariance exist, we 
define our terminology in the following. We call an operation rotationally 
invariant, if rotating its input yields an equally rotated output. Thus, 
rotation and operation are interchangeable. This notion follows the classical 
definition of rotation invariance for differential operators. 
Note that some recent CNN literature refers to this concept as  
equivariance.

\subsection{Our Contributions}
We translate PDE and variational models into their corresponding neural 
architectures and identify how they achieve rotation invariance. We start 
with simple two-dimensional diffusion models for greyscale images. Extending 
the connection \cite{APWS21,RH20,ZS20} between explicit schemes for these 
models and residual networks~\cite{HZRS16} (ResNets) leads to neural activation 
functions which couple network channels. Their result is based on a 
rotationally invariant measure involving specific channels representing 
differential operators. 

By exploring multi-channel and multiscale diffusion models, we generalise the 
concept of coupling to ResNeXt~\cite{XGDT17} architectures as an extension of 
the ResNet. Activations which couple all network channels preserve 
rotation invariance, but allow to design anisotropic models with a 
directional filtering. 

We derive three central design principles for rotationally invariant neural 
network design, discuss their effects on practical CNNs and evaluate their 
effectiveness within an experimental evaluation.
Our findings transfer inherent PDE concepts to CNNs and thus help to pave the 
way to more model-based and mathematically well-founded learning. 

\subsection{Related Work}
Several works connect numerical solution strategies for PDEs to CNN 
architectures \cite{APWS21,LHL20,LZLD18,OPF19,ZCF19} to obtain novel 
architectures with better performance or provable mathematical guarantees. 
Others are concerned with using neural networks to 
solve~\cite{CP20,HJE18,RPK19} or learn PDEs from 
data~\cite{LLD19,RBPK17,Sch17}. Moreover, the approximation capabilities 
\cite{DDFH21,GKNV21,KPRS21,TG19} and stability aspects 
\cite{APWS21,BRRS21,HR17,RDF20,RH20,TRPW20,ZS20} of CNNs are often analysed 
from a PDE viewpoint. 

The connections between neural networks and variational methods have become a 
topic of intensive research. The idea of learning the regulariser in a 
variational framework has gained considerable traction and brought the 
performance of variational models to a new level 
\cite{FDR21,LSO18,MDSL21,PHAP21,REM17}. The closely related idea of 
unrolling~\cite{MLE21,SABE19} the steps of a minimising algorithm for a 
variational energy and learning its parameters has been equally prominent and 
successful~\cite{AO17,AMOS19,BGKR19,CP16,HHNP20,KEKP20,KKHP17}. 

We exploit and extend connections between variational models
and diffusion processes~\cite{SchW98}, and their relations to residual 
networks~\cite{APWS21,RH20}. In contrast to our previous works 
\cite{APWS21,AWP20} which focussed on the one-dimensional setting and 
corresponding numerical algorithms, we now concentrate on two-dimensional 
diffusion models that incorporate different strategies to achieve rotation 
invariance. This allows us to transfer concepts of rotation invariance from 
PDEs to CNNs, which yield hitherto unexplored CNN design strategies. 

A simple option to learn a rotationally invariant model is to perform data 
augmentation~\cite{SSP03}, where the network is trained on randomly rotated 
input data. This strategy, however, only approximates rotation invariance and 
is heavily dependent on the data at hand. 

An alternative is to design the filters themselves in a rotationally 
invariant way, e.g. by weight restriction~\cite{CLKW18}. However, the 
resulting rotation invariance is too fine-grained: The filters as 
the smallest network component are not oriented. Thus, the model is not able to 
perform a directional filtering.

Other works~\cite{FG06,LSBP16} create a set of rotated input images and apply 
filters with weight sharing to this set. Depending on the amount of sampled 
orientations, this can lead to large computational overhead. 

An elegant solution for inherent rotation invariance is based on symmetry 
groups. Gens and Domingos~\cite{GD14} as well as Dieleman et al.~\cite{DDK16} 
propose to consider sets of feature maps which are rotated versions of each 
other. This comes at a high memory cost as four times as many feature maps need 
to be processed. Marcos et al.~\cite{MVT16} propose to rotate the filters 
instead of the features, with an additional pooling of orientations. However, 
the pooling reduces the directional information too quickly. A crucial downside 
of all these approaches is that they only use four orientations. This only 
yields a coarse approximation of rotation invariance. 

This idea has been generalized to arbitrary symmetry groups by Cohen and 
Welling~\cite{CW16} through the use of group convolution layers. Group 
convolutions lift the standard convolution to other symmetry groups which can 
also include rotations, thus leading to rotation invariance by design. 
However, also there, only four rotations are considered. This is remedied by 
Weiler et al.~\cite{WC19,WHS18} who make use of steerable 
filters~\cite{FA91} to design a larger set of oriented filters. Duits et 
al.~\cite{DSBP21} go one step further by formulating all layers as solvers to 
parametrised PDEs. Similar ideas have been implemented with 
wavelets~\cite{SM13} and circular harmonics \cite{WGTB17}, and the group 
invariance concept has also been extended to higher dimensional 
data~\cite{CGKW18,PRPO19,WGWB18}. However, processing multiple orientations in 
dedicated network channels inflates the network architecture, and discretising 
the large set of oriented filters requires the use of large stencils.

We provide an alternative by means of a more sophisticated activation 
function design. By coupling specific network channels, we can achieve inherent
rotation invariance without using large stencils or group theory, while still 
allowing for models to perform directional filtering. In a similar manner, 
Mr\'azek and Weickert proposed to design rotationally invariant wavelet 
shrinkage \cite{MW03} by using a coupling wavelet shrinkage function. However, 
to the best of our knowledge coupling activation functions have not been 
considered in CNNs so far.

\subsection{Organisation of the Paper}
We motivate our view on rotationally invariant design with a tutorial example 
in Section \ref{sec:motivation}. Afterwards, we review variational models and 
residual networks as the two other basic concepts in Section \ref{sec:review}. 
In Section \ref{sec:ours}, we connect various diffusion models and their 
associated energies to their neural counterparts and identify central concepts 
for rotation invariance. We summarise our findings and discuss their practical 
implementation in Section \ref{sec:discussion} and conduct experiments on 
rotation invariance in Section \ref{sec:experiments}. We finish the paper with 
our conclusions in Section \ref{sec:conclusions}. 

\section{Two Views on Rotational Invariance}\label{sec:motivation}
To motivate our viewpoint on rotationally invariant model design, 
we review a nonlinear diffusion filter of Weickert \cite{We94a} for image 
denoising and enhancement. It achieves anisotropy by integrating 
one-dimensional diffusion processes over all directions. This integration model 
creates a family of greyscale images $u(\bm x, t): \Omega \times [0, \infty) 
\rightarrow \mathbb{R}$ on an image domain $\Omega \subset \mathbb{R}^2$ 
according to the integrodifferential equation
\begin{equation}\label{eq:weickert_ecmi}
  \partial_t u = \frac{2}{\pi} \int_{0}^{\pi}
    \partial_{e_\theta} \left(g\!\left(\left|\partial_{e_\theta} 
    u_\sigma\right|^2\right) \partial_{e_\theta} u\right) \, d\theta,
\end{equation}
where $\partial_{e_\theta}$ is a directional derivative along the orientation 
of an angle $\theta$. The evolution is initialised as $u(\cdot,0) = f$ 
with the original image $f$, and reflecting boundary conditions are imposed.  
The model integrates one-dimensional nonlinear diffusion processes 
with different orientations $\theta$. All of them share a 
nonlinear decreasing diffusivity function $g$ which steers the diffusion in 
dependence of the local directional image structure $\left|\partial_{e_\theta} 
u_\sigma\right|^2$. Here, $u_\sigma$ is a smoothed version of $u$ which has 
been convolved with a Gaussian of standard deviation~$\sigma$. 

As this model diffuses more along low contrast directions than along high 
contrast ones, it is anisotropic. It is still rotationally invariant, since it 
combines all orientations of the one-dimensional processes with equal 
importance. However, this concept comes at the cost of an elaborate 
discretisation. First, one requires a large amount of discrete rotation angles 
for a reasonable approximation of the integration. Discretising the directional 
derivatives in all these directions with a sufficient order of consistency 
requires the use of large filter stencils; cf.~also \cite{BWC21}. The design of 
rotationally invariant networks such as~\cite{WHS18} faces similar 
difficulties. Processing the input by applying several rotated versions 
of an oriented filter requires large stencils and many orientations.

A much simpler option arises when considering the closely related 
edge-enhancing diffusion (EED) model \cite{We94e}
\begin{equation}\label{eq:divform}
  \partial_t u = \gradT\!\left(\bm D\!\left(\bm \nabla u_\sigma\right)\grad 
  u\right),
\end{equation}
where $\bm\nabla = \left(\partial_x, \partial_y\right)^\top$ denotes the 
gradient operator, and $\bm \nabla^\top$ is the divergence. Instead of an 
integration, the right hand side is given in divergence from. Thus, the 
process is now steered by a diffusion tensor $\bm D\! 
\left(\bm \nabla u_\sigma\right)$. It is a $2\times2$ positive semi-definite 
matrix which is designed to adapt the diffusion process to local directional 
information by smoothing along, but not across dominant image structures. This 
is achieved by constructing $\bm D$ from its normalised eigenvectors $\bm v_1 
\parallel \bm \nabla u_\sigma$ and $\bm v_2 \bot \bm \nabla u_\sigma$ which 
point across and along local structures. The corresponding eigenvalues 
$\lambda_1 = g\!\left(\left|\bm \nabla u_\sigma\right|^2\right)$ and 
$\lambda_2=1$ inhibit diffusion across dominant structures, and allow smoothing 
along them. Thus, the diffusion tensor can be written as  
\begin{equation}
  \bm D \!\left(\bm \nabla u_\sigma\right) = g\!\left(\left|\bm \nabla 
  u_\sigma\right|^2\right) \bm v_1 \bm v_1^\top + 1 \, \bm v_2 \bm v_2^\top. 
\end{equation}

Discretising the EED model~\eqref{eq:divform} is much more 
convenient. For example, a discretisation of the divergence term  with good 
rotation invariance can be performed on a $3\times3$ stencil, which is the 
minimal size for a consistent discretisation of a second order 
model~\cite{WWW13}. 

This illustrates a central insight: \emph{One can replace a complex 
discretisation by a sophisticated design of the nonlinearity}. This motivates 
us to investigate how rotationally invariant design principles 
of diffusion models translate into novel activation function designs. 

\section{Review: Variational Methods and Residual Networks}\label{sec:review}
We now briefly review variational methods and residual networks as the other 
two central concepts in our work. 

\subsection{Variational Regularisation}
Variational regularisation~\cite{Ti63,Wh23} obtains a function 
$u(\bm x)$ on a domain $\Omega$ as the minimiser of an 
energy functional. A general form of such a functional reads
\begin{equation}\label{eq:reggeneral}
  E(u) = \int_\Omega \left(D\!\left(u,f\right) + \alpha 
  R\!\left(u\right)\right) \, d\bm x.
\end{equation}
Therein, a data term $D(u,f)$ drives the solution $u$ to be close to an 
input image $f$, and a regularisation term $R(u)$ enforces smoothness 
conditions on the solution. The balance between the terms is controlled by a 
positive smoothness parameter $\alpha$.

We restrict ourselves to energy functionals with only a regularisation term and 
interpret the gradient descent to the energy as a parabolic diffusion PDE. This 
connection serves as one foundation for our findings. The variational framework 
is the simplest setting for analysing invariance properties, as these are 
automatically transferred to the corresponding diffusion process. 

\subsection{Residual Networks}
Residual networks (ResNets)~\cite{HZRS16} belong to the most popular neural 
network architectures to date. Their specific structure facilitates the 
training of very deep networks, and shares a close connection to PDE models. 

ResNets consist of chained residual blocks. A single residual block computes a 
discrete output $\bm u$ from an input $\bm f$ by means of 
\begin{equation}\label{eq:resblock}
  \bm u = \varphi_2\!\left(\bm f + \bm W_2\, \varphi_1\!\left(\bm W_1 \bm f + 
  \bm 
  b_1\right) + \bm b_2\right).
\end{equation}
First, one applies an inner convolution to $\bm f$, which is modelled by a 
convolution matrix $\bm W_1$. In addition, one adds a bias vector $\bm b_1$. 
The result of this inner convolution is fed into an inner \emph{activation} 
function $\varphi_1$. Often, these activations are fixed to simple 
functions such as the rectified linear unit (ReLU)~\cite{NH10} which is a 
truncated linear function:
\begin{equation}
  \text{ReLU}(s) = \text{max}(0,s).
\end{equation} 
The activated result is convolved with an outer convolution $\bm W_2$ with a 
bias vector $\bm b_2$. Crucially, the result of this convolution is added back 
to the original input signal $\bm f$. This \emph{skip connection} is the key to 
the success of ResNets, as it avoids the vanishing gradient phenomenon found in 
deep feed-forward networks~\cite{BSF94,HZRS16}. Lastly, one applies an outer 
activation function $\varphi_2$ to obtain the output $\bm u$ of the residual 
block. 

In contrast to diffusion processes and variational methods, these networks are 
not committed to a specific input dimensionality. In standard networks, the 
input is quickly deconstructed into multiple channels, each one concerned with 
different, specific image features. Each channel is activated independently, 
and information is exchanged through trainable convolutions. While this makes 
networks flexible, it does not take into account concepts such as rotation
invariance. By translating rotationally invariant diffusion models into ResNets 
and extensions thereof, we will see that shifting the focus from the 
convolutions towards activations can serve as an alternative way to guarantee 
built-in rotation invariance within a network.

\section{From Diffusion PDEs and Variational Models to Rotationally Invariant 
Networks}\label{sec:ours}

With the concepts from Sections~\ref{sec:motivation} and~\ref{sec:review}, we 
are now in a position to derive diffusion-inspired principles of rotationally 
invariant network design.

\subsection{Isotropic Diffusion on Greyscale Images}
We first consider the simplest setting of isotropic diffusion models 
for images with a single channel. By reviewing three popular models, we 
identify the common concepts for rotation invariance, and find a unifying 
neural network interpretation. 

We start with the second order diffusion model of Perona and Malik 
\cite{PM90}, which is given by the PDE
\begin{equation}\label{eq:pmsc}
\partial_t u = \gradT \left(g\!\left(\left|\grad u\right|^2\right) \grad 
u\right),
\end{equation}
with reflecting boundary conditions. This model creates a family of gradually 
simplified images $u(\bm x, t)$ according to the diffusivity $g(s^2)$. It 
attenuates the diffusion at locations where the gradient magnitude of the 
evolving image is large. In contrast to the model of Weickert 
\eqref{eq:weickert_ecmi}, the Perona--Malik model is isotropic, i.e. it does 
not have a preferred direction. 

The variational counterpart of this model helps us to identify the cause of its 
rotation invariance. An energy for the Perona--Malik model can be written in 
the following way which allows different generalisations:
\begin{equation}\label{eq:energy_pm}
  E(u) = \int_{\Omega} \Psi\!\left(\text{tr}\!\left(\grad u 
  \grad u^\top\right)\right) \, d\bm x,
\end{equation}
with an increasing regulariser function $\Psi$ which can be connected to the 
diffusivity $g$ by $g=\Psi^\prime$ \cite{SchW98}. Comparing the 
functional~\ref{eq:energy_pm} to the one in~\eqref{eq:reggeneral}, we have now 
specified the form of the regulariser to be $R(u) 
=\Psi\!\left(\text{tr}\!\left(\grad u \grad u^\top\right)\right)$.

The argument of the regulariser is the trace of the so-called structure 
tensor~\cite{FG87}, here without Gaussian regularisation, which reads
\begin{equation}\label{eq:structensor}
  \grad u \grad u^\top = \begin{pmatrix}
  u_{x}^2 & 
  u_{x}u_{y}\\ 
  u_{x}u_{y}
  & u_{y}^2
  \end{pmatrix}.
\end{equation}
This structure tensor is a $2 \times 2$ matrix with eigenvectors $\bm v_1 
\parallel \grad u$ and $\bm v_2 \bot \grad u$ parallel and orthogonal to the 
image gradient. The corresponding eigenvalues are given by $\nu_1 = \left|\grad 
u\right|^2$ and $\nu_2 = 0$, respectively. Thus, the eigenvectors span a local 
coordinate system where the axes point across and along dominant structures of 
the image, and the larger eigenvalue describes the magnitude of image 
structures.

The use of the structure tensor is the key to rotation invariance. A rotation 
of the image induces a corresponding rotation of the structure tensor and the 
structural information that it encodes: Its eigenvectors rotate along, and its 
eigenvalues remain unchanged. Consequently, the trace as the sum of the 
eigenvalues is rotationally invariant.

In the following, we explore other ways to design the energy functional based 
on rotationally invariant quantities and investigate how the resulting 
diffusion model changes.

The fourth order model of You and Kaveh~\cite{YK00} relies on the Hessian 
matrix. The corresponding energy functional reads
\begin{equation}\label{eq:youkaveh_energy}
E(u) = \int_{\Omega} \Psi\!\left(\left(\text{tr}\!\left(\bm 
H(u)\right)\right)^2\right) \, 
d\bm x.
\end{equation}
Here, the regulariser takes the squared trace of the Hessian matrix $\bm H(u)$ 
as an argument. Since the trace of the Hessian is equivalent to the Laplacian 
$\Delta u$, the gradient flow of~\eqref{eq:youkaveh_energy} can be written as
\begin{equation}\label{eq:youkaveh}
  \partial_t u = - \Delta \left(g\!\left(\left(\Delta 
      u\right)^2\right) \Delta u\right).
\end{equation}
This is a fourth order counterpart to the Perona--Malik model. Instead of the 
gradient operator, one considers the Laplacian $\Delta$. This change was 
motivated as one remedy to the staircasing effect of the Perona--Malik model 
\cite{YK00}.

The rotationally invariant matrix at hand is the Hessian $\bm H(u)$. In 
a similar manner as the structure tensor, the Hessian describes local structure 
and thus follows a rotation of this structure. Also in this case, the trace 
operation reduces the Hessian to a scalar that does not change under rotations.

To avoid speckle artefacts of the model of You and Kaveh, Lysaker et 
al.~\cite{LLT03} propose to combine all entries of the Hessian in the 
regulariser. They choose the Frobenius norm of the Hessian $||\bm 
H(u)||^2_F$ together with a total variation regulariser. For more general 
regularisers, this model reads~\cite{DWB09}
\begin{equation}
  E(u) = \int_{\Omega} \Psi\!\left(||\bm H(u)||^2_F\right) \, d\bm x,
\end{equation}
which yields a diffusion process of the form 
\begin{equation}\label{eq:lysaker}
  \partial_t u = - \mathcal{D}^\top \!\left(g\!\left(||\bm H(u)||^2_F\right) 
                   \mathcal{D} u\right),
\end{equation}
where the differential operator $\mathcal D$ induced by the Frobenius norm reads
\begin{equation}
  \mathcal D = \left(\partial_{xx}, \partial_{xy}, \partial_{yx},
                 \partial_{yy}\right)^\top.
\end{equation}
This shows another option how one can use the rotationally invariant 
information of the Hessian matrix. While the choice of a Frobenius norm instead 
of the trace operator changes the associated differential operators in the 
diffusion model, it does not destroy the rotation invariance property: The 
squared Frobenius norm is the sum of the squared eigenvalues of the Hessian, 
which in turn are rotationally invariant. 

\subsection{Coupled Activations for Operator Channels}
In the following, we extend the connections between residual networks and 
explicit schemes from~\cite{APWS21,RH20,ZS20} in order to transfer rotation 
invariance concepts to neural networks. To this end, we consider the 
generalised diffusion PDE
\begin{equation}\label{eq:gen_iso}
  \partial_t u = - \mathcal{D}^* \! \left(g\!\left(|\mathcal D u|^2\right) 
  \mathcal D u\right).
\end{equation}
Here, we use a generalised differential operator $\mathcal D$ and its adjoint 
$\mathcal D^*$. This PDE subsumes the diffusion 
models~\eqref{eq:pmsc},~\eqref{eq:youkaveh}, and~\eqref{eq:lysaker}.
Since the diffusivities take a scalar argument, we can express 
the diffusivity as $g(|\mathcal D u|^2)$. The differential operator $\mathcal 
D$ is induced by the associated energy functional. 

To connect the generalised model~\eqref{eq:gen_iso} to a ResNet architecture, 
we first rewrite~\eqref{eq:gen_iso} by means of the vector-valued flux function 
$\bm \Phi(\bm s) = g(|\bm s|^2)\, \bm s$ as 
\begin{equation}\label{eq:fluxform}
  \partial_t u = - \mathcal{D}^* \left(\bm \Phi\!\left(\mathcal{D} 
  u\right)\right).
\end{equation}
Let us now consider an explicit discretisation for this diffusion PDE. 
The temporal derivative is discretised by a forward difference with time step 
size $\tau$, and the spatial derivative operator $\mathcal{D}$ is discretised 
by a convolution matrix $\bm K$. Consequently, the adjoint $\mathcal{D}^*$ is 
discretised by $\bm K^\top$. Depending on the number of components of 
$\mathcal{D}$, the matrix $\bm K$ implements a set of convolutions. This yields 
an explicit scheme for \eqref{eq:fluxform}
\begin{equation}\label{eq:explicit}
  \bm u^{k+1} = \bm u^k - \tau \bm K^\top \bm \Phi\!\left(\bm K \bm 
    u^k\right).
\end{equation}
where a superscript $k$ denotes the discrete time level. One can connect this 
explicit step~\eqref{eq:explicit} to a residual block~\eqref{eq:resblock} by 
identifying 
\begin{equation}\label{eq:resnet_relation}
  \bm W_1 = \bm K, \quad 
  \varphi_1 = \tau \bm \Phi, \quad 
  \bm W_2 = -\bm K^\top, \quad
  \varphi_2 = \text{Id},
\end{equation}
and setting the bias vectors to $\bm 0$~\cite{APWS21,RH20,ZS20}.

In contrast to the one-dimensional considerations in \cite{APWS21}, the  
connection between flux function and activation in the two-dimensional setting 
yields additional, novel design concepts for activation functions. This yields 
the first design principle for neural networks. 

\begin{designprinciple}[Coupled Activations for Rotational 
Invariance]\label{dp:coupling}
  Activation functions which couple network channels can be used to design 
  rotationally invariant networks. At each position of the image, the channels 
  of the inner convolution result are combined within a rotationally invariant 
  quantity which determines the nonlinear response.  
\end{designprinciple}

The coupling effect of the diffusivity and the regulariser directly transfers 
to the activation function. This is apparent when the differential operator 
$\mathcal D$ contains multiple components. For example, consider an operator 
$\mathcal D = \left(\mathcal D_1, \mathcal D_2\right)^\top$ with two components 
and its discrete variant $\bm K = \left(\bm K_1, \bm K_2\right)^\top$. 

The application of the operator $\bm K$ transforms the single-channel 
signal~$\bm u^k$ into a signal with two channels. Then the activation function 
couples the information from both channels within the diffusivity $g$. For each 
pixel position~$i,j$, we have 
\begin{equation}
  \bm \Phi 
    \begin{pmatrix}
    \left(\bm K_1 \bm u^k\right)_{i,j} \\ \left(\bm K_2 \bm 
    u^k\right)_{i,j}
    \end{pmatrix}
  = g\left( \left|\bm K_1 \bm u^k\right|_{i,j}^2 + \left|\bm K_2 \bm 
  u^k\right|_{i,j}^2  \right)
    \begin{pmatrix}
    \left(\bm K_1 \bm u^k\right)_{i,j} \\ \left(\bm K_2 \bm 
        u^k\right)_{i,j}
    \end{pmatrix}.
\end{equation} 
Afterwards, the application of $\bm K^\top$ reduces the resulting two-channel 
signal to a single channel again.

In the general case, the underlying differential operator $\mathcal D$ 
determines how many channels are coupled. The choice $\mathcal D = 
\left(\partial_{xx}, \partial_{xy}, \partial_{yx}, \partial_{yy}\right)^\top$ 
of Lysaker et al.~\cite{LLT03} induces a coupling of four channels containing 
second order derivatives. This shows that a central 
condition for rotation invariance is that the convolution $\bm K$ implements 
a rotationally invariant differential operator. We discuss the effects of this 
condition on the practical filter design in Section~\ref{sec:discussion}.

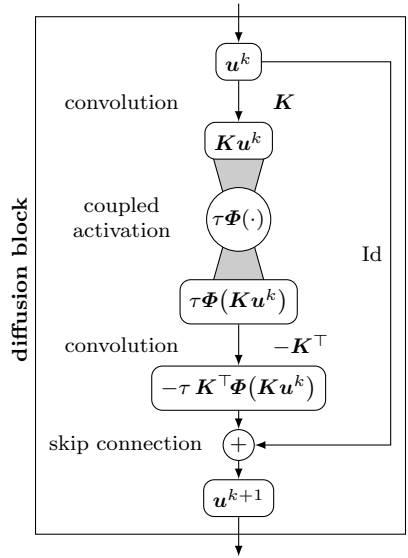
\begin{figure}
  \centering
  \pgfdeclarelayer{bg0}
\pgfdeclarelayer{bg1}
\pgfsetlayers{bg0,bg1,main}
\begin{tikzpicture}[-latex]
  \matrix (m)
  [
    matrix of nodes,
    column 1/.style = {nodes={block}}
  ]
  {
    |(input)|          
    $\bm u^k$
    \\[4.5ex]            
    |(fwd)|
    $\bm K \bm u^k$%
    \\[3ex]
    |[operator] (act)|
    $\tau \bm \Phi\!\left(\cdot\right)$%
    \\[3ex]
    |(flux)|
    $\tau \bm\Phi\!\left(\bm K \bm u^k\right)$%
    \\[4.5ex]
    |(bwd)|
    $-\tau \, \bm K^\top \!\bm\Phi\!\left(\bm K \bm u^k\right)$%
    \\[2ex]
    |[operator] (add)|
    $+$%
    \\[2ex]
    |(output)|         
    $\bm u^{k+1}$%
    \\
  };

    \def\nameshift{-1.8em}                    
    \def\skipconnshift{2.0}                     

    \layername{input}{fwd}{convolution}{\nameshift}   
    \edge{input}{fwd}
    \edgelabel{$\bm K$}
    
    \coupleedgein{fwd}{act}     
    \coupleedgeout{act}{flux}     
    \layername{fwd}{flux}{\makecell{coupled \\activation}}{\nameshift}         
                                              
    \layername{flux}{bwd}{convolution}{\nameshift}                             
    \edge{flux}{bwd}          
    \edgelabel{$-\bm K^\top$}

    \node[left of=add, xshift=-4ex] {skip connection};
    \edge{bwd}{add}                             

    \edge{add}{output}
  
    \skipedge{input}{add}{\skipconnshift}   

    \node[above of=input, yshift=-1ex] (TMP) {};
    \draw (TMP) -- (input);

    \node[below of=output, yshift=1ex] (TMP) {};
    \draw (output) -- (TMP);

    \node[left of=input, xshift=\nameshift-3.8em, yshift=2em] (topleft) {};
    \node[right of=output, xshift=4em, yshift=-1.6em] (bottomright) {};
    \draw (topleft) rectangle (bottomright);

    \node [left of=flux, rotate=90, anchor=north, yshift=7em, xshift=1em]
          {\textbf{diffusion block}};
\end{tikzpicture}
  \caption{Diffusion block for an explicit diffusion   
    step~\eqref{eq:explicit} with activation function $\tau\bm\Phi$, 
    time step size $\tau$, and convolution filters $\bm K$. The activation 
    function couples the channels of the operator $\bm K$.
    \label{fig:block_singlechannel}}
\end{figure}

We call a block of the form~\eqref{eq:explicit} a \emph{diffusion 
block}. It is visualized in Figure~\ref{fig:block_singlechannel} in graph form. 
Nodes contain the state of the signal, while edges describe the operations to 
move from one state to another. We denote the channel coupling by a shaded 
connection to the activation function. 

The coupling effect is natural in the diffusion case. However, to the 
best of our knowledge, this concept has not been proposed for CNNs in the 
context of rotation invariance.

\subsection{Diffusion on Multi-channel Images}
So far, the presented models have been isotropic. They only consider the 
magnitude of local image structures, but not their direction. However, we will 
see that anisotropic models inspire another form of activation function which 
combines directional filtering with rotation invariance. 

To this end, we move to diffusion on multi-channel images. While there are 
anisotropic models for single-channel images~\cite{We97}, they require a 
presmoothing as shown in the EED model~\eqref{eq:divform}. However, 
such models do not have a conventional energy formulation~\cite{We21}. The 
multi-channel setting allows one to design anisotropic models that do not 
require a presmoothing and arise from a variational energy.

In the following we consider multi-channel images $\bm u = \left(u_1, u_2, 
\dots, u_M\right)^\top$ with $M$ channels. To distinguish them from the 
previously considered channels of the differential operator, we refer to image 
channels and operator channels in the following. 

A naive extension of the Perona--Malik model \eqref{eq:pmsc} to multi-channel 
images would treat each image channel separately. Consequently, the energy 
would consider a regularisation of the trace of the structure tensor for each 
individual channel. This in turn does not respect the fact that structural 
information is correlated in the channels. 

To incorporate this correlation, Gerig et al.~\cite{GKKJ92} 
proposed to sum up structural information from all channels. An 
energy functional for this model reads
\begin{equation}\label{eq:energy_isomc}
 E(\bm u) = \bigintss_{\Omega} \Psi\!\left(\text{tr }\sum_{m=1}^{M}\grad 
    u_m \grad u_m^\top\right) \, d\bm x.
\end{equation}
Here, we again use the trace formulation. It shows that this model 
makes use of a colour structure tensor, which goes back to Di 
Zenzo~\cite{Di86}. It is the sum of the structure tensors of the individual 
channels. In contrast to the single-channel structure tensor without Gaussian 
regularisation, no closed form solution for its eigenvalues and eigenvectors 
are available. Still, the sum of structure tensors stays rotationally 
invariant. 

The corresponding diffusion process is described by the coupled 
PDE set
\begin{equation}\label{eq:gerig}
  \partial_t u_m = \gradT \left(g\!\left(\sum_{n=1}^{M}\left|\grad 
    u_n\right|^2\right) \grad u_m\right)
  \qquad \left(m=1,\dots,M\right),
\end{equation}
with reflecting boundary conditions. As trace and summation are 
interchangeable, the argument of the regulariser corresponds to a sum of 
channel-wise gradient magnitudes. Thus, the diffusivity considers 
information from all channels. It allows to steer the diffusion process in all 
channels depending on a joint structure measure. 

Interestingly, a simple change in the energy model~\eqref{eq:energy_isomc} 
incorporates directional information~\cite{WB02} such that the model becomes 
anisotropic. Switching the trace operator and the regulariser yields the energy 
\begin{equation}\label{eq:aniso_energy}
E(\bm u) = \bigintss_{\Omega} 
    \text{tr }\Psi\!\left(\sum_{m=1}^M\grad 
    u_m \grad u_m^\top\right) \, d\bm x.
\end{equation}
Now the regulariser acts on the colour structure tensor in the sense of a power 
series. Thus, the regulariser modifies the eigenvalues $\nu_1, \nu_2$ to 
$\Psi\!\left(\nu_1\right), \Psi\!\left(\nu_2\right)$ and leaves the 
eigenvectors unchanged. For the $2\times2$ colour structure tensor we have 
\begin{equation}
  \Psi\!\left(\sum_{m=1}^M\grad u_m \grad u_m^\top\right)
  = \Psi\!\left(\nu_1\right) \bm v_1 \bm v_1^\top +
    \Psi\!\left(\nu_2\right) \bm v_2 \bm v_2^\top.
\end{equation}
The eigenvalues are treated individually. This allows for an anisotropic 
model, as each eigenvalue determines the local image contrast along its 
corresponding eigenvector. Still, the model is rotationally invariant as the 
colour structure tensor rotates accordingly. Consequently, the trace of this 
regulariser is equivalent to the sum of the regularised eigenvalues:
\begin{equation}
 \text{tr }\Psi\!\left(\sum_{m=1}^M\grad u_m \grad u_m^\top\right)
  = \Psi\!\left(\nu_1\right)  + \Psi\!\left(\nu_2\right).
\end{equation}
This illustrates the crucial difference to the isotropic case, where we have 
\begin{equation}
  \Psi\!\left(\text{tr }\sum_{m=1}^M\grad u_m \grad u_m^\top\right)
  = \Psi\!\left(\nu_1 + \nu_2\right),
\end{equation}
Both eigenvalues of the structure tensor are regularised jointly and the result 
is a scalar, which shows that no directional information can be involved. 
At this point, the motivation for using the structure tensor notation in the 
previous models becomes apparent: Switching the trace operator and the 
regulariser changes an isotropic model into an anisotropic one. 

The gradient descent of the energy~\eqref{eq:aniso_energy} is an anisotropic 
nonlinear diffusion model for multi-channel images~\cite{WB02}:
\begin{equation}\label{eq:eed_mc}
  \partial_t u_m = \gradT \left(g\!\left(\sum_{n=1}^M\grad 
        u_n \grad u_n^\top\right) \grad u_m\right)
          \qquad \left(m=1,\dots,M\right).
\end{equation}
The diffusivity inherits the matrix-valued argument of the regulariser. Thus, 
it is applied in the same way and yields a $2\times2$ diffusion tensor. In 
contrast to single-channel diffusion, this creates anisotropy as its 
eigenvectors do not necessarily coincide with $\grad u$. Thus, the 
multi-channel case does not require Gaussian presmoothing.

We have seen that the coupling effect within the diffusivity goes beyond the 
channels of the differential operator. It combines both the operator channels 
as well as the image channels within a joint measure. Whether the model is 
isotropic or anisotropic is determined by the shape of the diffusivity result: 
Isotropic models use scalar diffusivities, while anisotropic models require 
matrix-valued diffusion tensors. In the following, we generalise this concept 
and analyse its influence on the ResNet architecture. 

\subsection{Coupled Activations for Image Channels}
A generalised formulation of the multichannel diffusion models~\eqref{eq:gerig} 
and~\eqref{eq:eed_mc} is given by
\begin{equation}
  \partial_t u_m = - \mathcal{D}^* 
  \bm\Phi\!\left(\bm u, \mathcal{D} u_m\right)
    \qquad \left(m=1,\dots,M\right).
\end{equation}
As the flux function uses more information than only $\mathcal{D} u_m$, we 
switch to the notation $\bm\Phi\!\left(\bm u, \mathcal{D} u_m\right)$. 
An explicit scheme for this model is derived in a similar way as before, 
yielding
\begin{equation}\label{eq:coupled_mc}
  \bm u_m^{k+1} = \bm u_m^k - \tau \bm K^\top \bm \Phi \! \left(\bm u^k, \bm K 
  \bm u_m^k\right)  \qquad \left(m=1,\dots,M\right).
\end{equation}
The activation function now couples more than just the operator channels, it 
couples all its input channels. In contrast to Design 
Principle~\ref{dp:coupling}, this coupling is more general and provides a 
second design principle. 

\begin{designprinciple}[Fully Coupled Activations for Image 
Channels]\label{dp:fullcoupling}
  Activations which couple both operator channels and image channels can be 
  used to create anisotropic, rotationally invariant models. At each position 
  of the image, all operator channels for all image channels are combined 
  within a rotationally invariant quantity which determines the nonlinear 
  response.
\end{designprinciple}

The different coupling effects serve different purposes: Coupling the 
image channels accounts for structural correlations and can be used to create 
anisotropy. Coupling the channels of the differential operators guarantees 
rotation invariance. 

This design principle becomes apparent when explicitly formulating the 
activation function. Isotropic models use a scalar diffusivity within the flux 
function
\begin{equation}
  \bm \Phi \! \left(\bm u^k, \left(\bm K \bm u_m^k\right)_{i,j}\right)  = 
  g\!\left(\sum_{m=1}^M\left|\bm K \bm u^k_m\right|^2_{i,j}\right) \left(\bm K 
  \bm u^k_m\right)_{i,j}
  \qquad \left(m=1,\dots,M\right),
\end{equation}
which couples all channels of $\bm u$ at the position $i,j$, as well as all 
components of the discrete operator $\bm K$. Anisotropic models require a 
matrix-valued diffusion tensor in the flux function
\begin{equation}
  \bm \Phi \! \left(\bm u^k \!, \!\left(\bm K \bm u_m^k\right)_{i,j}\right)  = 
  g\!\left(\sum_{m=1}^M \! \left(\bm K \bm u^k_m\right)_{i,j} \!\left(\bm K \bm 
  u^k_m\right)_{i,j}^\top\right) \! \left(\bm K \bm u^k_m\right)_{i,j}
  \,\, \left(m=1,\dots,M\right),
\end{equation}

\begin{figure}
  \centering
  \resizebox{\linewidth}{!}{\pgfdeclarelayer{bg0}
\pgfdeclarelayer{bg1}
\pgfsetlayers{bg0,bg1,main}
\begin{tikzpicture}[-latex]
  \matrix (m)
  [
    matrix of nodes,
    column 1/.style = {nodes={block}},
    column 2/.style = {nodes={block}},
    column 3/.style = {nodes={block}},
    column 4/.style = {nodes={block}},
    column 5/.style = {nodes={block}},
    row 6 column 3/.style={overlay}, 
    row 8 column 3/.style={overlay}, 
  ]
  {
      %
    &[3ex]  
      %
    &[-7ex] 
      |(input)|
      $\bm u_1^k, \dots, \bm u_M^k$%
    &[0ex]  
      %
    &[0ex]  
      %
    \\[7ex] 
      |(grad1)|
      $ \bm K \bm u_1^k$%
    &
      |(grad2)|
      $ \bm K \bm u_2^k$%
    &
      %
    &
      |[draw=none]|
      \large $\cdots$%
    &
      |(gradC)|
      $ \bm K  \bm u_M^k$%
    \\[3ex] 
          %
        &
          %
        &
         |[operator] (act)|
          $\tau \bm \Phi\!\left(\bm u^k, \cdot\right)$%
        &
          %
        &
          %
    \\[3ex] 
      |(flux1)|
      $ \tau   \bm\Phi\! \left(\bm u^k, \bm K \bm u_1^k\right)$%
    &
      |(flux2)|
      $ \tau   \bm\Phi\! \left(\bm u^k, \bm K \bm u_2^k\right)$%
    &
      %
    &
      |[draw=none]|
      \large $\cdots$%
    &
      |(fluxC)|
      $ \tau   \bm\Phi\! \left(\bm u^k, \bm K \bm u_M^k\right)$%
    \\[7ex] 
      |(div1)|
      $ -\tau   \bm K^\top \bm\Phi\! \left(\bm u^k, \bm K \bm 
      u_1^k\right)$%
    &
      |(div2)|
      $ -\tau   \bm K^\top \bm\Phi\! \left(\bm u^k, \bm K \bm 
      u_2^k\right)$%
    &
      %
    &
      |[draw=none]|
      \large $\cdots$%
    &
      |(divC)|
      $ -\tau   \bm K^\top \bm\Phi\! \left(\bm u^k, \bm K \bm 
      u_M^k\right)$%
    \\[7ex] 
      %
    &
      %
    &
      |(concat)|
      $-\tau   \bm K^\top \bm\Phi\! \left(\bm u^k, \bm K \bm 
            u_1^k\right), \dots, -\tau   \bm K^\top \bm\Phi\! \left(\bm u^k, 
            \bm K \bm u_M^k\right)$%
    &
      %
    &
      %
    \\[7ex] 
      %
    &
      %
    &
      |[operator] (add)|
      $+$%
    &
      %
    &
      %
    \\[4ex] 
      %
    &
      %
    &
      |(output)|
      $\bm u_1^{k+1}, \dots, \bm u_M^{k+1}$%
    &
      %
    &
      %
    \\
  };

    \def\nameshift{-3em}                    
    \def\skipconnshift{4.8}                   

    \layernamesplit{input}{grad1}{convolution}{\nameshift}
    \splitedge{input}{grad1}
    \edgelabel{$\bm K$}
    \splitedge{input}{grad2}
    \edgelabel{$\bm K$}
    \splitedge{input}{gradC}
    \edgelabel{$\bm K$}
  
    \coupleedgein{grad1}{act}
    \coupleedgeout{act}{flux1}
    \coupleedgein{grad2}{act}
    \coupleedgeout{act}{flux2}
    \coupleedgein{gradC}{act}
    \coupleedgeout{act}{fluxC}
    \layername{grad1}{flux1}{\makecell{fully coupled \\ 
    activation}}{\nameshift}         
    
    \layername{flux1}{div1}{convolution}{\nameshift}
    \edge{flux1}{div1}
    \edgelabel{$-\bm K^\top$}
    \edge{flux2}{div2}
    \edgelabel{$-\bm K^\top$}
    \edge{fluxC}{divC}
    \edgelabel{$-\bm K^\top$}
  
    \joinedge{div1}{concat}
    \joinedge{div2}{concat}
    \joinedge{divC}{concat}
  
    \edge{concat}{add}
  
    \edge{add}{output}
  
    \skipedge{input}{add}{\skipconnshift}   
    \node[left of=add, xshift=-18em] {skip connection};
  
    \node[above of=input, yshift=0] (TMP) {};
    \draw (TMP) -- (input);
    
    \node[below of=output, yshift=0] (TMP) {};
    \draw (output) -- (TMP);
  
    \node[left of=input, xshift=\nameshift-18.8em, yshift=2em] (topleft) {};
    \node[right of=output, xshift=13.5em, yshift=-1.6em] (bottomright) {};
    \draw (topleft) rectangle (bottomright);
  
   \node [left of=grad1, rotate=90, anchor=north, xshift=-8em, yshift=8em]
          {\textbf{fully coupled multi-channel diffusion block}};
\end{tikzpicture}}
   \caption{Fully coupled multi-channel diffusion block for an explicit 
            step~\eqref{eq:coupled_mc} with a fully coupled activation function 
            $\tau\bm\Phi$, time step size $\tau$, and convolution filters $\bm 
            K$. The activation function couples all operator and image channels 
            of its input jointly. Depending on the design of the activation, 
            the resulting model can be isotropic or anisotropic.   
            \label{fig:block_multichannel_fullcoupling}}
\end{figure}
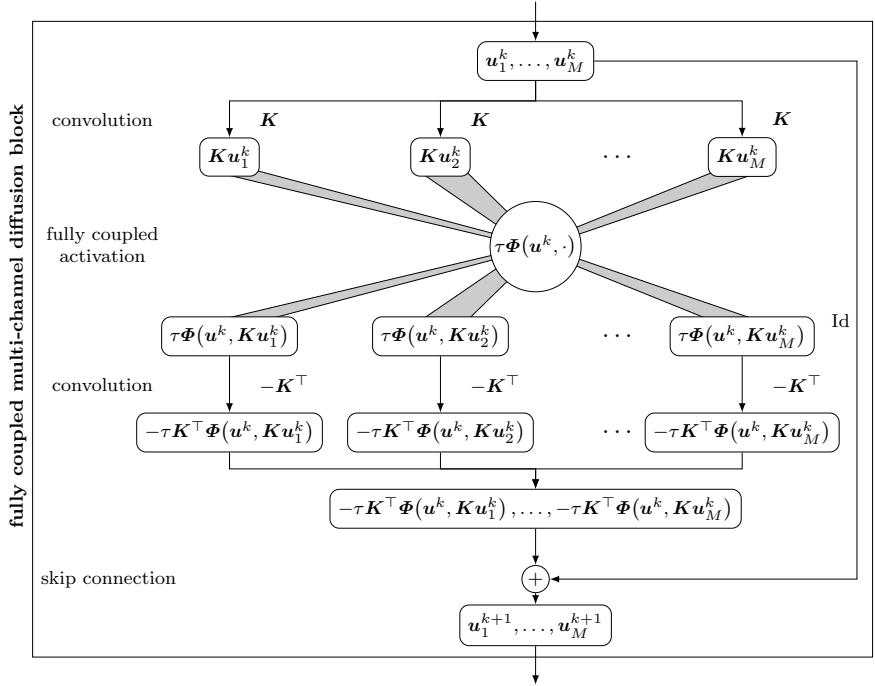

This concept is visualised in Figure \ref{fig:block_multichannel_fullcoupling} 
in the form of a \emph{fully coupled multi-channel diffusion block}. To clarify 
the distinction between image and operator channels, we explicitly split 
the image into its channels. We see that all information of the inner filter 
passes through a single activation function and influences all outgoing results 
in the same manner. 

Design Principle~\ref{dp:fullcoupling} shows that coupling cannot only be used 
for rotationally invariant design, but also makes sense for implementing 
modelling aspects such as anisotropy. This is desirable as anisotropic models 
often exhibit higher performance through better adaptivity to data. 

\subsection{Integrodifferential Diffusion}
The previous models work on the finest scale of the image. However, image 
structures live on different scales of the image. Large image structures are 
present on coarser scales than fine ones. Generating a structural measure which 
incorporates information from multiple image scales can be beneficial. 

To this end, we consider integrodifferential extensions of single scale 
diffusion which have proven advantageous in practical applications such as 
denoising~\cite{AW21}. In analogy to the multi-channel diffusion setting, these 
models inspire a full coupling of scale information for a variation of residual 
networks.

We start with the energy functional
\begin{equation}
  E\!\left(u\right) = \int_\Omega
        \Psi\!\left(\text{tr }\int_{0}^{\infty} \left(\mathcal D^{(\sigma)} 
        u\right) \left(\mathcal D^{(\sigma)} u\right)^\top
        d\sigma \right) \,   d\bm{x}.
\end{equation}
We denote the scale parameter by $\sigma$ and assume that the differential 
operators $\mathcal D^{(\sigma)}$ are dependent on the scale. This can be 
realised for example by an adaptive presmoothing of an underlying differential 
operator; see e.g.~\cite{AW21,DW07}.

Instead of summing structure tensors over image channels, this model integrates 
generalised structure tensors $\left(\mathcal D^{(\sigma)} u\right) 
\left(\mathcal D^{(\sigma)} u\right)^\top$ over multiple scales. This results 
in a multiscale structure tensor~\cite{AW21} which contains a semi-local 
measure for image structure. If $\mathcal D^{(\sigma)}$ are rotationally
invariant operators, then the multiscale structure tensor is also invariant. 

The corresponding diffusion model reads
\begin{equation}\label{eq:iso_ms}
  \partial_t u = - \int_{0}^{\infty}\mathcal{D}^{(\sigma)*} 
    \left(g\!\left(\int_{0}^{\infty}\left|\mathcal D^{(\gamma)} 
      u\right|^2 \, d\gamma \right)\mathcal{D}^{(\sigma)} u \right) 
      \, d\sigma,
\end{equation}
where $g = \Psi^\prime$. Due to the chain rule, one obtains two integrations 
over the scales: The outer integration combines diffusion processes on each 
scale. The inner integration, where the scale variable has been renamed to 
$\gamma$, accumulates multiscale information within the diffusivity argument. 

This model is a variant of the integrodifferential isotropic diffusion model of 
Alt and Weickert~\cite{AW21}. Therein, however, the diffusivity uses a 
scale-adaptive contrast parameter. Thus, it does not arise from an energy 
functional. 

As in the multi-channel diffusion models, switching trace and regulariser 
yields an anisotropic model, which is described by the energy
\begin{equation}
 E\!\left(u\right) = \int_\Omega
      \text{tr }\Psi\!\left(\int_{0}^{\infty}\left(\mathcal D^{(\sigma)} 
      u\right) \left(\mathcal D^{(\sigma)} u\right)^\top d\sigma \right) \, 
      d\bm{x}.
\end{equation}
In analogy to the multi-channel model, the regulariser is applied directly to 
the structure tensor, which creates anisotropy.
Consequently, the resulting diffusion process is a variant of the 
integrodifferential anisotropic diffusion~\cite{AW21}: 
\begin{equation}\label{eq:aniso_ms}
  \partial_t u = -\!\int_{0}^{\infty}\!\mathcal{D}^{(\sigma)*} 
       \left(g\!\left(\int_{0}^{\infty}\left(\mathcal{D}^{(\gamma)} u\right) 
       \left(\mathcal  D^{(\gamma)} u\right)^\top d\gamma \right) 
       \mathcal{D}^{(\sigma)} 
        u\right) d\sigma.
\end{equation}
The anisotropic regularisation is inherited by the diffusivity and results in a 
flux function that implements a matrix-vector multiplication.

\subsection{Coupled Activations for Image Scales}
Both the isotropic and the anisotropic multiscale models can be summarised by 
the flux formulation
\begin{equation}\label{eq:multiscale_generalised}
  \partial_t u = - \int_{0}^{\infty}\mathcal{D}^{(\sigma)*} 
    \left(\bm \Phi\!\left(u, \mathcal{D}^{(\sigma)} u\right) \right)
      \, d\sigma.
\end{equation}
To discretise this model, we now require a discretisation of the scale 
integral. To this end, we select a set of $L$ discrete scales $\sigma_1, 
\sigma_2, \dots, \sigma_L$. On each scale~$\sigma_\ell$, we employ discrete 
differential operators $\bm K_\ell$.
This yields an explicit scheme for the continuous 
model~\eqref{eq:multiscale_generalised} which reads
\begin{equation}\label{eq:fullcoupling}
  \bm u^{k+1} = \bm u^k  - \tau \sum_{\ell=1}^{L}\omega_\ell \, \bm K_\ell^\top 
    \bm \Phi \! \left(\bm u^k, \bm K_\ell \bm u^k\right).
\end{equation}
Here, $\omega_\ell$ is a step size over the scales, discretising the 
infinitesimal quantity~$d\sigma$. It is dependent on the scale to allow a 
non-uniform sampling of scales $\sigma_\ell$. A simple choice is $\omega_\ell = 
\sigma_{\ell+1} - \sigma_{\ell}$.

Interestingly, an extension of residual networks called ResNeXt~\cite{XGDT17} 
provides the corresponding neural architecture to this model.
Therein, the authors consider a sum of transformations of the input 
signal together with a skip connection. We restrict ourselves to 
the following formulation:
\begin{equation}\label{eq:resnext}
  \bm u = \varphi_2\left(\bm f + \sum_{\ell = 1}^{L} \bm W_{2,\ell} \, 
              \varphi_\ell \!
              \left(\bm W_{1,\ell} \bm f + \bm b_{1,\ell}\right) + \bm 
              b_{2,\ell}\right).
\end{equation}
This ResNeXt block modifies the input image $\bm f$ within $L$ 
independent paths, and sums up the results before the skip connection. Each 
path may apply multiple, differently shaped convolutions. Choosing a single 
path with $L=1$ yields the ResNet model. 

We can identify an explicit multiscale diffusion step~\eqref{eq:fullcoupling} 
with a ResNeXt block~\eqref{eq:resnext} by
\begin{equation}
  \bm W_{1,\ell} = \bm K_\ell, \quad 
  \varphi_{1,\ell} = \tau \, \bm \Phi, \quad 
  \bm W_{2,\ell} = -\omega_\ell\bm K^\top_\ell, \quad
  \varphi_2 = \text{Id},
\end{equation}
and all bias vectors $\bm b_{1,\ell}, \bm b_{2,\ell}$ are set to $\bm 0$, for 
all $\ell=1,\dots,L$.

In contrast to the previous ResNet relation \eqref{eq:resnet_relation}, we 
apply different filters~$\bm K_\ell$ in each path. Their individual results are 
summed up before the skip connection, which approximates the scale integration. 
While the ResNeXt block allows for individual activation functions in each 
path, we use a common activation with a full coupling for all of them. This 
constitutes a variant of Design Principle~\ref{dp:fullcoupling}, where one now 
couples image scales.

\begin{figure}
  \centering
  \resizebox{\linewidth}{!}{\pgfdeclarelayer{bg0}
\pgfdeclarelayer{bg1}
\pgfsetlayers{bg0,bg1,main}
\begin{tikzpicture}[-latex]
  \matrix (m)
  [
    matrix of nodes,
    column 1/.style = {nodes={block}},
    column 2/.style = {nodes={block}},
    column 3/.style = {nodes={block}},
    column 4/.style = {nodes={block}},
    column 5/.style = {nodes={block}},
    row 7 column 3/.style={overlay},
  ]
  {
      %
    &[3ex]  
      %
    &[-7ex] 
      |(input)|
      $\bm u^k$%
    &[0ex]  
      %
    &[0ex]  
      %
    \\[7ex] 
      |(grad1)|
      $ \bm K_1 \bm u^k$%
    &
      |(grad2)|
      $ \bm K_2 \bm u^k$%
    &
      %
    &
      |[draw=none]|
      \large $\cdots$%
    &
      |(gradL)|
      $ \bm K_L  \bm u^k$%
    \\[5ex] 
      %
    &
      %
    &
     |[operator,minimum width=1cm] (act)|
      $\tau \bm \Phi\!\left(\bm u^k, \cdot\right)$%
    &
      %
    &
      %
    \\[5ex] 
      |(flux1)|
      $ \tau \, \bm\Phi\! \left(\bm u^k, \bm K_1 \bm u^k\right)$%
    &
      |(flux2)|
      $ \tau \, \bm\Phi\! \left(\bm u^k, \bm K_2 \bm u^k\right)$%
    &
      %
    &
      |[draw=none]|
      \large $\cdots$%
    &
      |(fluxL)|
      $ \tau \, \bm\Phi\! \left(\bm u^k, \bm K_L \bm u^k\right)$%
    \\[7ex] 
      |(div1)|
      $ -\tau \, \bm K_1^\top \bm\Phi\! \left(\bm u^k, \bm K_1 \bm u^k\right)$%
    &
      |(div2)|
      $ -\tau \, \bm K_2^\top \bm\Phi\! \left(\bm u^k, \bm K_2 \bm u^k\right)$%
    &
      %
    &
      |[draw=none]|
      \large $\cdots$%
    &
      |(divL)|
      $ -\tau \, \bm K_L^\top \bm\Phi\! \left(\bm u^k, \bm K_L \bm u^k\right)$%
    \\[7ex] 
          %
        &
          %
        &
          |[operator] (pathadd)|
          $+$%
        &
          %
        &
          %
        \\[4ex] 
          %
        &
          %
        &
          |(sum)|
          $- \tau \sum_{\ell=1}^{L}\omega_\ell\,\bm K_\ell^\top 
            \bm \Phi_\ell\!\left(\bm u^k,\bm K_\ell \bm u^k\right)$
        &
          %
        &
          %
        \\[4ex] 
          %
        &
          %
        &
          |[operator](add)|
          $+$%
        &
          %
        &
          %
        \\[4ex] 
          %
        &
          %
        &
          |(output)|
          $\bm u^{k+1}$%
        &
          %
        &
          %
        \\
  };

    \def\nameshift{-3em}                    
    \def\skipconnshift{4.8}                   

    \layernamesplit{input}{grad1}{convolution}{\nameshift}
    \splitedge{input}{grad1}
    \edgelabel{$\bm K_1$}
    \splitedge{input}{grad2}
    \edgelabel{$\bm K_2$}
    \splitedge{input}{gradL}
    \edgelabel{$\bm K_L$}
  
    \coupleedgein{grad1}{act}
    \coupleedgeout{act}{flux1}
    \coupleedgein{grad2}{act}
    \coupleedgeout{act}{flux2}
    \coupleedgein{gradL}{act}
    \coupleedgeout{act}{fluxL}
    \layername{grad1}{flux1}{\makecell{fully coupled \\ 
    activation}}{\nameshift}         
    
    \layername{flux1}{div1}{convolution}{\nameshift}
    \edge{flux1}{div1}
    \edgelabel{$-\bm K_1^\top$}
    \edge{flux2}{div2}
    \edgelabel{$-\bm K_2^\top$}
    \edge{fluxL}{divL}
    \edgelabel{$-\bm K_L^\top$}
  
    \joinedge{div1}{pathadd}
    \edgelabel{$\cdot\,\omega_1$}
    \joinedge{div2}{pathadd}
    \edgelabel{$\cdot\,\omega_2$}
    \joinedge{divL}{pathadd}
    \edgelabel{$\cdot\,\omega_L$}
  
    \edge{pathadd}{sum}
    \node[left of=pathadd, xshift=-18.5em] {\makecell{weighted sum \\over 
    scales}};
    \edge{sum}{add}
    \edge{add}{output}
  
    \skipedge{input}{add}{\skipconnshift}   
    \node[left of=add, xshift=-18.5em] {skip connection};
    
    \node[above of=input, yshift=0] (TMP) {};
    \draw (TMP) -- (input);
    
    \node[below of=output, yshift=0] (TMP) {};
    \draw (output) -- (TMP);
  
    \node[left of=input, xshift=\nameshift-20em, yshift=2em] (topleft) {};
    \node[right of=output, xshift=14em, yshift=-1.6em] (bottomright) {};
    \draw (topleft) rectangle (bottomright);
  
   \node [left of=grad1, rotate=90, anchor=north, xshift=-10em, yshift=8.7em]
          {\textbf{fully coupled multiscale diffusion block}};
\end{tikzpicture}}
  \caption{Fully coupled multiscale diffusion block for an explicit multiscale 
    diffusion step~\eqref{eq:fullcoupling} with a single activation function 
    $\tau\omega_\ell\bm\Phi$, time step size $\tau$, and convolution filters 
    $\bm K_\ell$ on each scale. The activation function couples all inputs 
    jointly. Depending on the design of the activation, the resulting model 
    can be isotropic or anisotropic. 
  \label{fig:block_multiscale_fullcoupling}}
\end{figure}
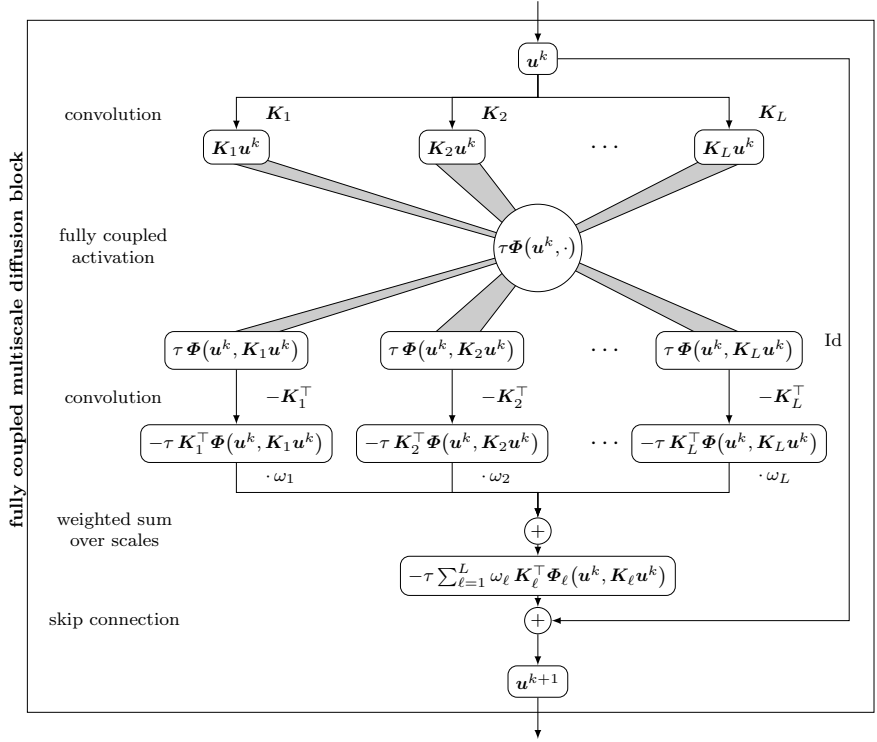

\begin{designprinciple}[Fully Coupled Activations for Image Scales]
  \label{dp:fullcouplingscales}
  Activations which couple both operator channels and image scales can be 
  used to create anisotropic, rotationally invariant multiscale models. 
  At each position of the image, all operator channels for all image scales
  are combined within a rotationally invariant quantity which determines the 
  nonlinear response.
\end{designprinciple}

Also in this case, the combined coupling serves different purposes. Coupling 
the operator channels yields rotation invariance, and coupling of scales 
allows to obtain a more global representation of the image structure. 
Isotropic models employ a coupling with a scalar diffusivity in the flux 
function
\begin{equation}
  \bm \Phi\!\left(\bm u^k, \left(\bm K_\ell \bm u^k\right)_{i,j}\right) 
  = g\!\left(\sum_{\ell=1}^L\left|\bm K_\ell
        \bm u^k\right|_{i,j}^2 \right)
        \left(\bm K_\ell \bm u^k\right)_{i,j},
\end{equation}
and a matrix-valued diffusion tensor in the flux function
\begin{equation}
  \bm \Phi\!\left(\bm u^k, \left(\bm K_\ell \bm u^k\right)_{i,j}\right) 
  = g\!\left(\sum_{\ell=1}^L\left(\bm K_\ell
        \bm u^k\right)_{i,j} \left(\bm K_\ell
                \bm u^k\right)_{i,j}^\top\right)
        \left(\bm K_\ell \bm u^k\right)_{i,j}
\end{equation}
can be used to create anisotropic models. 

We call a block of the form~\eqref{eq:fullcoupling} a \emph{fully coupled 
multiscale diffusion block}. This block is visualised in 
Figure~\ref{fig:block_multiscale_fullcoupling}.
Comparing its form to that of the multichannel diffusion block in 
Figure~\ref{fig:block_multichannel_fullcoupling}, one can see that the 
different architectures use the same activation function design, however 
with different motivations.

\afterpage{
\begin{landscape}
\begin{table}
  \setlength{\tabcolsep}{.8mm}
  \everymath{\displaystyle}
  \caption{The considered diffusion models, along with their variational 
  energies and the resulting network architectures.  
  \label{tab:models}}
  \resizebox{\linewidth}{!}{ 
  \begin{tabular}{|c|c|c|c|c|}
    \hline\TstrutS
    model 
    & variational energy 
    & diffusion PDE
    & explicit scheme / network block
    & \makecell{activation design}  
    \BstrutS\\ \hline\TstrutL
    \makecell{Perona--Malik~\cite{PM90},\\ single-channel,\\ isotropic}
    & $E(u) = \int_{\Omega} \Psi\!\left(\text{tr}\!\left(\grad u \grad 
       u^\top\right)\right) \, d\bm x$
    & $\partial_t u = \gradT \left(g\!\left(\left|\grad u\right|^2\right) \grad 
    u\right)$
    & \multirow{9}*{$ \bm u^{k+1} = \bm u^k - \tau \bm K^\top \bm 
    \Phi\!\left(\bm K \bm u^k\right)$} 
    & \makecell{isotropic coupling via \\ structure tensor,\\ scalar 
    multiplication}
    \BstrutL\\\cline{1-3}\cline{5-5}\TstrutL
    \makecell{You and Kaveh~\cite{YK00},\\ single-channel,\\ isotropic}
    & $E(u) = \int_{\Omega} \Psi\!\left(\left(\text{tr}\!\left(\bm 
    H(u)\right)\right)^2\right) \, d\bm x$
    & $\partial_t u = - \Delta \left(g\!\left(\left(\Delta 
    u\right)^2\right) \Delta u\right)$
    &
    & \makecell{isotropic coupling via Hessian,\\ scalar multiplication}
    \BstrutL\\\cline{1-3}\cline{5-5}\TstrutL
    \makecell{Lysaker et al.~\cite{LLT03},\\ single-channel,\\ isotropic}
    & $E(u) = \int_{\Omega} \Psi\!\left(||\bm H(u)||^2_F\right) \, d\bm x$
    & \makecell{$\partial_t u = - \mathcal{D}^*
                             \left(g\!\left(||\bm H(u)||^2_F\right) 
                             \mathcal{D}
                                   u\right)$
     \\[1mm] with $\mathcal D = \left(\partial_{xx},
                                         \partial_{xy},
                                         \partial_{yx},
                                         \partial_{yy}\right)^\top$}
    &
    & \makecell{isotropic coupling via Hessian,\\ scalar multiplication}
    \BstrutL\\\hline\TstrutL
    \makecell{Gerig et al.~\cite{GKKJ92},\\ coupled multi--channel,\\ 
    isotropic}
    & $E(\bm u) = \bigintssss_{\Omega} \Psi\!\left(\text{tr }\sum_{m=1}^M\grad 
    u_m \grad u_m^\top\right) \, d\bm x$
    & $\partial_t u_m = \gradT \left(g\!\left(\sum_{n=1}^M\left|\grad 
    u_n\right|^2\right) \grad u_m\right)$
    & \multirow{6}*{$\bm u_m^{k+1} = \bm u_m^k - \tau \bm K^\top \bm 
    \Phi\!\left(\bm u^k, \bm K \bm  u_m^k\right) $} 
    & \makecell{isotropic coupling via\\ multi-channel structure tensor,\\ 
    scalar multiplication}
    \BstrutL\\\cline{1-3}\cline{5-5}\TstrutL
    \makecell{Weickert and Brox~\cite{WB02},\\ coupled multi--channel,\\ 
    anisotropic}
    & $E(\bm u) = \bigintssss_{\Omega} 
    \text{tr }\Psi\!\left(\sum_{m=1}^M\grad 
    u_m \grad u_m^\top\right) \, d\bm x$
    & $\partial_t u_m = \gradT \left(g\!\left(\sum_{n=1}^M\grad 
        u_n \grad u_n^\top\right) \grad u_m\right)$
    &
    & \makecell{ansotropic coupling via\\ multi-channel structure tensor,\\ 
    matrix-vector multiplication}
    \BstrutL\\\hline\TstrutL
    \makecell{Alt and Weickert~\cite{AW21},\\ coupled multiscale,\\ 
    isotropic}
    &\resizebox{0.5\textwidth}{!}{$E\!\left(u\right)\!=\! \int_\Omega \!
            \Psi\!\left(\text{tr}\int_{0}^{\infty}\!\! \left(\mathcal 
            D^{(\sigma)} 
            u\right)\! \left(\mathcal D^{(\sigma)} u\right)^\top \!
            \!d\sigma \right)\! d\bm{x}$}
    &\resizebox{0.5\textwidth}{!}{$\partial_t u 
    -\!\int_{0}^{\infty}\!\!\mathcal{D}^{(\sigma)*} 
       \left(g\!\left(\int_{0}^{\infty}\!\left|\mathcal D^{(\gamma)} 
       u\right|^2  d\gamma \right) \mathcal{D}^{(\sigma)} 
        u \right) d\sigma$}
    & \multirow{6}*{\resizebox{0.4\textwidth}{!}{$ \bm u^{k+1} = \bm u^k  - 
    \tau \sum_{\ell=1}^{L}\omega_\ell \, \bm K_\ell^\top 
    \bm \Phi \! \left(\bm u^k, \bm K_\ell \bm u^k\right)$}} 
    & \makecell{isotropic coupling via\\ multiscale structure tensor,\\    
                scalar multiplication}
    \BstrutL\\\cline{1-3}\cline{5-5}\TstrutL
    \makecell{Alt and Weickert~\cite{AW21},\\ coupled multiscale,\\ 
    anisotropic}
    &\resizebox{0.5\textwidth}{!}{$E\!\left(u\right)\!=\!  \int_\Omega
      \text{tr}\,\Psi\!\left(\int_{0}^{\infty}\!\!\left(\mathcal D^{(\sigma)} 
      u\right)\! \left(\mathcal D^{(\sigma)} u\right)^\top \!d\sigma \right) \, 
      \!d\bm{x}$}
    &\resizebox{0.5\textwidth}{!}{$\partial_t u\! =\! 
    -\!\int_{0}^{\infty}\!\!\mathcal{D}^{(\sigma)*} 
       \!\left(g\!\left(\int_{0}^{\infty}\!\!\left(\mathcal{D}^{(\gamma)} 
       u\right) \!\!
       \left(\mathcal  D^{(\gamma)} u\right)^{\!\top} \!\!d\gamma 
       \right) 
       \mathcal{D}^{(\sigma)} 
        u \right) d\sigma$}
    &
    & \makecell{anisotropic coupling via\\ multiscale structure tensor,\\
     matrix-vector multiplication}
    \BstrutL\\\hline
  \end{tabular}}
\end{table}
\end{landscape}
}
\section{Discussion} \label{sec:discussion}
We have seen that shifting the design focus from convolutions to activation 
functions can yield new insights into CNN design. We summarise all models that 
we have considered in Table \ref{tab:models} as a convenient overview. 

All variational models are rotationally invariant, as they rely on a structural 
measure which accounts for rotations. This directly transfers to the diffusion 
model, its explicit scheme, and thus also its network counterpart, resulting in 
Design Principle~\ref{dp:coupling}. Moreover, the different coupling options 
for models with multiple scales and multiple channels show how a sophisticated 
activation design can steer the model capacity. This has led to the 
additional Design Principles~\ref{dp:fullcoupling} 
and~\ref{dp:fullcouplingscales}.

The coupling effects are naturally motivated for diffusion, but are 
hitherto unexplored in the CNN world. While activation functions such as 
maxout~\cite{GWMC13} and softmax introduce a coupling of their input arguments, 
they only serve the purpose of reducing channel information. Even though some 
works focus on using trainable and more advanced 
activations~\cite{CP16,FQXC17,OMLM18}, the coupling aspect has not been 
considered so far.

The rotation invariance of the proposed architectures can be 
approximated efficiently in the discrete setting. For example for second order 
models, Weickert et al.~\cite{WWW13} present $L^2$ stable discretisations with 
good practical rotation invariance that only require a $3\times3$ stencil. 
For models of second order, this is the smallest possible discretisation 
stencil which still yields consistent results. 

In a practical setting with trainable filters, one is not restricted to the 
differential operators that we have encountered so far. To guarantee that the 
learned filter corresponds to a rotationally invariant differential operator, 
one has several options. For example, one can design the filters based on a 
dictionary of operators which fulfil the rotation invariance property, which 
are then combined into more complex operators through trainable weights. In a 
similar manner, one can employ different versions of a base operator which 
arise from a rotationally invariant operation, e.g. a Gaussian smoothing. We 
pursue this strategy in our experiments in the following section in analogy to 
\cite{AW21}. 

Apart from the coupling aspect, the underlying network architecture is not 
modified. This is a stark contrast to the CNN literature where a set of 
orientations is discretised, requiring much larger stencils for a good 
approximation of rotation invariance. We neither require involved 
discretisations, nor a complicated lifting to groups. Thus, we regard the 
proposed activation function design as a promising alternative to the 
directional splitting idea. 

\section{Experiments} \label{sec:experiments}
In the following, we present an experimental evaluation to support our 
theoretical considerations. To this end, we design trainable multiscale 
diffusion models for denoising. We compare models with and without coupling 
activations, and evaluate their performance on differently rotated datasets. 
This shows that the Design Principle~\ref{dp:coupling} is indeed necessary for 
rotation invariance. 

\subsection{Experimental Setup}

We train the isotropic and anisotropic multiscale diffusion 
models~\eqref{eq:iso_ms} and \eqref{eq:aniso_ms}. Both perform a full coupling 
of all scales, i.e. they implement Design Principles~\ref{dp:coupling} 
and~\ref{dp:fullcouplingscales}. As a counterpart, we train the same 
multiscale diffusion model with the diffusivity applied to each channel of the 
discrete derivative operator separately. Thus, the activation is applied 
independently in each direction. This violates Design 
Principle~\ref{dp:coupling}. Hence, the model should yield worse rotation 
invariance than the coupled models. 

Still, all models implement Design Principle~\ref{dp:fullcouplingscales} by 
integrating multiscale information. For an evaluation of the importance of this 
design principle we refer to \cite{AW21}, where multiscale models outperform 
their single scale counterparts.

The corresponding explicit scheme for the considered models is given by 
\begin{equation}
\bm u^{k+1} = \bm u^k  - \tau \sum_{\ell=1}^{L}\omega_\ell \, \bm K_\ell^\top 
    \bm \Phi \! \left(\bm u^k, \bm K_\ell \bm u^k\right)
\end{equation}
The choice for $\omega_\ell$ is set to $\sigma_{\ell+1}-\sigma_{\ell}$.

As differential operators $\bm K_\ell$, we choose
weighted, Gaussian smoothed gradients $\beta_\ell \bm \nabla_{\sigma_\ell}$ on 
each scale $\sigma_\ell$. The application of a smoothed gradient to an image 
via $\bm \nabla_{\sigma} u = G_{\sigma} \ast \bm \nabla u$ is equivalent to 
computing a Gaussian convolution with standard deviation $\sigma$ of the image 
gradient. Moreover, we weight the differential operators on each scale by a  
scale-adaptive, trainable parameter~$\beta_\ell$. 

A discrete set of $L=8$ scales is determined by an exponential sampling between 
a minimum scale of $\sigma_{\text{min}}=0.1$ and a maximum one of 
$\sigma_{\text{max}}=10$. This yields discrete scales $[0.1, 
0.18,0.32,0.56,1.0,1.77,3.16,5.62]$.

To perform edge-preserving denoising, we choose the exponential 
Perona--Malik~\cite{PM90} diffusivity
\begin{equation}
  g(s^2) = \exp\left(-\frac{s^2}{2\lambda^2}\right). 
\end{equation}
It attenuates the diffusion at locations where the argument exceeds a contrast 
parameter $\lambda$. This parameter is trained in addition to the 
scale-adaptive weights. 

Moreover, we train the time step size $\tau$ and we use $10$ explicit steps 
with shared parameter sets. This amounts to a total number of $10$ trainable 
parameters: $\tau$, $\lambda$, and $\beta_1$ to $\beta_8$. 

In the practical setting, a discretisation with good rotation invariance is 
crucial. We use the nonstandard finite difference discretisation of Weickert et 
al.~\cite{WWW13}. It implements the discrete divergence term
\begin{equation}
  \bm A(\bm u^k) = \bm K_\ell^\top \bm \Phi \! \left(\bm u^k, \bm K_\ell \bm 
  u^k\right)
\end{equation}
on a stencil of size $3\times3$. For isotropic models, it has a free 
parameter $\alpha \in \left[0, \frac12\right]$ which can be tuned for rotation 
invariance, with an additional parameter $\gamma \in \left[0, 1\right]$ for 
anisotropic ones. We found that in the denoising case, the particular choice of 
these two parameters constitutes a trade-off between performance and rotation 
invariance.

We train all models on a synthetic dataset which consists of greyscale images 
of size $256\times256$ with values in the range $\left[0, 255\right]$. Each 
image contains~$20$ randomly placed white rectangles of size $140\times70$ on a 
black background. The rectangles are all oriented along a common direction, 
which creates a directional bias within the dataset. The training set contains 
$100$ images and is oriented with an angle of $30^\circ$ from the $x$-axis. As 
test datasets, we consider rotated versions of a similar set of $50$ images. 
The rotation angles are sampled between $0^\circ$ and $90^\circ$ in steps of 
$5^\circ$. To avoid an influence of the image sampling on the evaluation, we 
exclude the axis-aligned datasets.

To train the models for the denoising task, we add noise of standard deviation 
$60$ to the clean training images and minimise the Euclidean distance to 
the ground truth images. We measure the denoising quality in terms of 
peak-signal-to-noise ratio (PSNR). All models are trained for $250$ epochs with 
the Adam optimiser~\cite{KB14} with standard settings and a learning rate of 
$0.001$. One training epoch requires $50$ seconds on an \emph{NVIDIA 
GeForce GTX 1060 6GB}, and the evaluation on one of the test sets requires $7$ 
seconds.

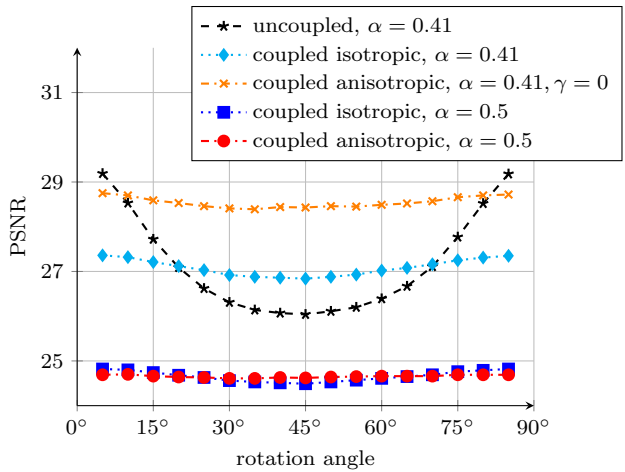
\begin{figure}
  \centering
  \begin{tikzpicture}[scale = 1.0]

\begin{axis}
[
width=0.9*\axisdefaultwidth,
height=0.75*\axisdefaultwidth,
samples=500,
axis lines=left,
ymin = 24, ymax = 32,
xmin = 0, xmax = 90,
xtick = {0,15, ..., 90},
xticklabels={$0^\circ$,$15^\circ$,$30^\circ$,$45^\circ$,
             $60^\circ$,$75^\circ$,$90^\circ$},
ytick = {25, 27, 29, 31},
ylabel={PSNR},
xlabel={rotation angle},
xlabel near ticks,
ylabel near ticks,
grid = major,
legend entries={uncoupled, $\alpha=0.41$\\
                coupled isotropic, $\alpha=0.41$\\
                coupled anisotropic, $\alpha=0.41, \gamma=0$\\
                coupled isotropic, $\alpha=0.5$\\
                coupled anisotropic, $\alpha=0.5$\\
                },
legend cell align=left,
legend style={at={(0.25, 0.9)}, anchor = west},
]

\addplot+[thick, dashed, mark options={solid}, black, mark = star]
table[x = degree, y = uncoupled]
{errors};

\addplot+[thick, dotted, mark options={solid}, cyan, mark = diamond*]
table[x = degree, y = isotropic041]
{errors};

\addplot+[thick, dashdotted, mark options={solid}, orange, mark = x]
table[x = degree, y = anisotropic041]
{errors};

\addplot+[thick, dotted, mark options={solid}, blue, mark = square*]
table[x = degree, y = isotropic05]
{errors};

\addplot+[thick, dashdotted, mark options={solid}, red, mark = *]
table[x = degree, y = anisotropic05]
{errors};

\end{axis}
\end{tikzpicture}
  \caption{Denoising quality on differently rotated versions of the test 
  dataset. The models have been trained on a dataset with $30^\circ$ 
  orientation. The models with coupling approximate rotation invariance 
  significantly better than the uncoupled model. 
  \label{fig:experiment}} 
\end{figure}

\begin{figure}
  \centering 
  \begin{tabular}{cc}
  ground truth
  &\makecell{noisy, \\ PSNR $15.6$ dB}
  \\
  \includegraphics[width=0.24\linewidth]{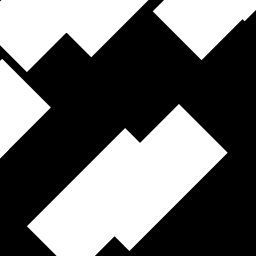}
  &\includegraphics[width=0.24\linewidth]{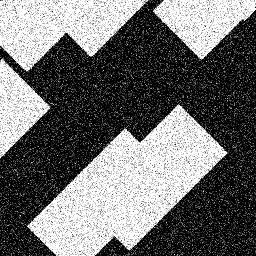}
  \end{tabular}
  \begin{tabular}{ccc}
  \includegraphics[width=0.24\linewidth]{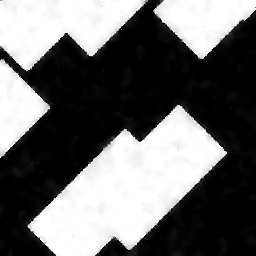}
  &\includegraphics[width=0.24\linewidth]{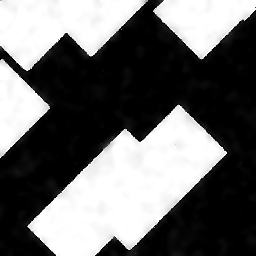}
  &\includegraphics[width=0.24\linewidth]{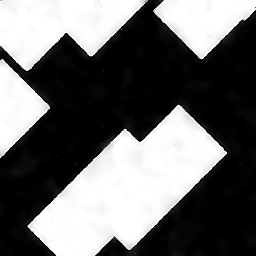}
  \\
  \makecell{uncoupled, \\ PSNR $27.0$ dB} 
  & \makecell{coupled isotropic, \\ PSNR $28.1$ dB} 
  & \makecell{coupled anisotropic, \\ PSNR $29.2$ dB} 
  \end{tabular}
  \caption{Visual comparison of denoising results for a rotation angle of 
  $45^\circ$. The coupled models achieve better denoising quality as they 
  generalise better to the rotated data. 
  \label{fig:visual}} 
\end{figure}

\subsection{Evaluation}

A rotationally invariant model should produce the same PSNR on all rotations of 
the test dataset. Thus, in Figure \ref{fig:experiment} we plot the PSNR on the 
test datasets against their respective rotation angle. 

We see that the fluctuations within both anisotropic and 
isotropic coupled models are much smaller than those within the uncoupled 
model. A choice of $\alpha=0.41$ and $\gamma=0$ yields a good balance 
between performance and rotation invariance. However, rotation 
invariance can also be driven to the extreme: A choice of $\alpha=0.5$, which 
renders the choice of $\gamma$ irrelevant, eliminates rotational fluctuations 
almost completely, but also drastically reduces the quality. The reason for 
this is given by Weickert et al.~\cite{WWW13}: A value of $\alpha=0.5$ 
decouples the image grid into two decoupled checkerboard grids which do not 
communicate except at the boundaries.

For the balanced choice of $\alpha=0.41$, the anisotropic model 
consistently outperforms the isotropic one, as it can smooth along oriented 
structures. As the uncoupled model can only do this for structures which are 
aligned with the $x$- and $y$-axes, it performs better the closer the rotation 
is to $0^\circ$ and $90^\circ$, respectively. Hence, it performs worst for a 
rotation angle of $45^\circ$. Thus, it does not achieve rotation invariance. 

We measure the rotation invariance in terms of the variance of the test 
errors over the rotation angles. While the isotropic and anisotropic coupled 
models with $\alpha=0.41$ achieve variances of $0.035$ dB and $0.014$ dB, 
the uncoupled model suffers from a variance of $1.25$ dB. The extreme choice of 
$\alpha=0.5$ even reduces the variances of the coupled models to $0.013$ dB and 
$8.7\cdot10^{-4}$~dB, respectively.

A visual inspection of the results in Figure~\ref{fig:visual} supports 
this trend. Therein, we present the denoised results on an example from the 
test data set with $45^\circ$ orientation. The uncoupled model suffers from 
ragged edges as the training on the differently rotated dataset has introduced 
a directional bias. The coupled isotropic model preserves the edges far better, 
and the coupled anisotropic model can even smooth along them to obtain the best 
reconstruction quality.

These findings show that the coupling effect leads 
to significantly better rotation invariance properties.

\section{Conclusions} \label{sec:conclusions}
We have seen that the connection between diffusion and neural networks allows 
to bring novel concepts for rotation invariance to the world of CNNs. The 
models which we considered inspire different activation function designs, which 
we summarise in Table \ref{tab:models}. 

The central design principle for rotation invariance is a coupling of 
operator channels. Diffusion models and their associated variational energies 
apply their respective nonlinear design functions to rotationally invariant 
quantities based on a coupling of multi-channel differential operators. Thus, 
the activation function as their neural counterpart should employ this 
coupling, too. Moreover, coupling image channels or scales in addition allows 
to create anisotropic models with better measures for structural information. 

This strategy provides an elegant and minimally invasive modification of 
standard architectures. Thus, coupling activation functions constitute a 
promising alternative to the popular network designs of splitting orientations 
and group methods in orientation space. Evaluating these concepts in practice 
and transferring them to more general neural network models is part of our 
ongoing work.

\bibliographystyle{spmpsci}       
\bibliography{myrefs}         

\end{document}